%% file: acl_latex.tex
\pdfoutput=1

\documentclass[11pt]{article}

\usepackage[]{acl}

\usepackage{times}
\usepackage{latexsym}

\usepackage[T1]{fontenc}

\usepackage[utf8]{inputenc}

\usepackage{microtype}

\usepackage{hyperref}
\usepackage{url}
\usepackage{graphicx}
\usepackage{tabularx}
\usepackage{mathtools}
\usepackage{booktabs}
\usepackage{tabu}
\usepackage{multirow}
\usepackage{arydshln}

\title{Take One Step at a Time to Know Incremental Utility of Demonstration:\\
       An Analysis on Reranking for Few-Shot In-Context Learning}

\author{Kazuma Hashimoto~~~~~~~~~Karthik Raman~~~~~~~~Michael Bendersky \\
{Google Research, Mountain View} \\
{\texttt{\{kazumah, karthikraman, bemike\}@google.com} }}

\begin{document}
\maketitle

\begin{abstract}
In-Context Learning (ICL) is an emergent capability of Large Language Models (LLMs).
Only a few demonstrations enable LLMs to be used as blackbox for new tasks.
Previous studies have shown that using LLMs' outputs as labels is effective in training models to select demonstrations.
Such a label is expected to estimate utility of a demonstration in ICL;
however, it has not been well understood how different labeling strategies affect results on target tasks.
This paper presents an analysis on different utility functions by focusing on LLMs' output probability given ground-truth output, and task-specific reward given LLMs' prediction.
Unlike the previous work, we introduce a novel labeling method, incremental utility, which estimates how much incremental knowledge is brought into the LLMs by a demonstration.
We conduct experiments with instruction-tuned LLMs on binary/multi-class classification, segmentation, and translation across Arabic, English, Finnish, Japanese, and Spanish.
Our results show that (1) the probability is effective when the probability values are distributed across the whole value range (on the classification tasks), and (2) the downstream metric is more robust when nuanced reward values are provided with long outputs (on the segmentation and translation tasks).
We then show that the proposed incremental utility further helps ICL by contrasting how the LLMs perform with and without the demonstrations.
\end{abstract}

\input{Sec-Introduction}

\input{Sec-Pipelines}

\input{Sec-Utility}

\input{Sec-Experiments}

\input{Sec-Related}

\input{Sec-Conclusion}

\input{Sec-Extra}

\bibliography{custom}
\bibliographystyle{acl_natbib}

\clearpage
\input{Sec-Appendix}

\end{document}

%% file: Sec-Introduction.tex
\section{Introduction}

Recent advances of Large Language Models (LLMs)~\cite{palm2,gpt4,llama2} have been pushing the field of Natural Language Processing (NLP) to the next level in many different aspects.
A notable capability of LLMs is few-shot {\it In-Context Learning (ICL)}, which uses only a few {\it demonstrations} (i.e., input-output pairs) to perform new tasks without finetuning~\cite{llms-few-shot,icl-bias-analysis}.

A crucial research topic is demonstration selection for ICL.
\citet{icl_retrieval} proposed retrieval-based ICL, and \citet{icl_retrieval_fine_1} have shown that further finetuning retrieval models is effective~\citep{icl_retrieval_fine_2,icl_retrieval_fine_3,icl_retrieval_fine_4}.
However, it is not conclusive how to train the selection models in several aspects: training labels, objectives, and overall pipelines.

\begin{figure}[t]
\includegraphics[width=\linewidth]{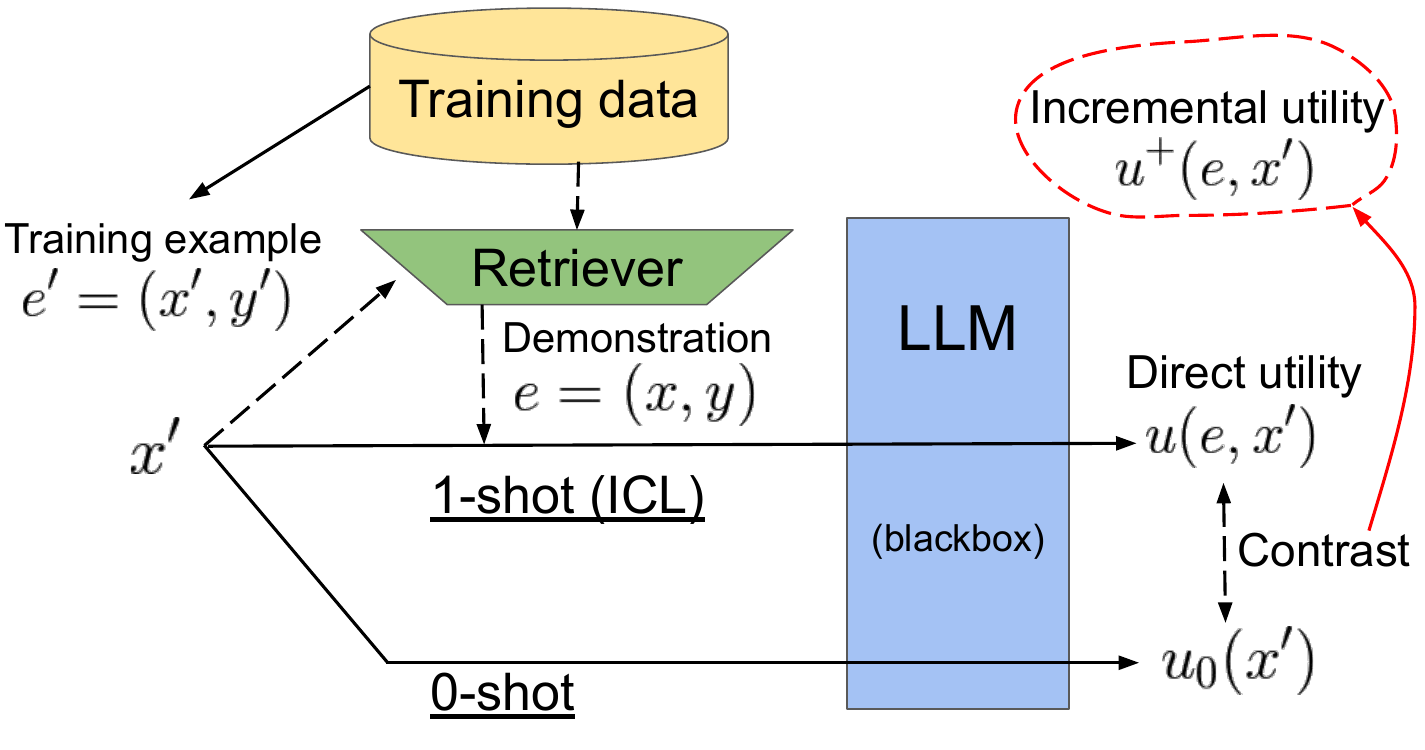}
\caption{Incremental utility of a demonstration.}
\label{fig:method}
\end{figure}

This paper focuses on the training labels of the selection models.
First, we use two types of values to estimate utility of a demonstration for an input: (1) an LLM's output probability of generating the ground-truth output (as in the previous work), and (2) a task-specific reward function given the LLM's prediction (as in reward optimization~\citep{seq2seq-rl}).
Next, we introduce a novel method to estimate {\it incremental utility} by contrasting the 1-shot ICL performance and the 0-shot performance (Figure~\ref{fig:method}).
We expect the incremental utility to estimate how much incremental knowledge is brought by the demonstration, assuming that the LLMs have a capability of 0-shot prediction~\citep{llms-few-shot}.

We conduct experiments in a standard retrieval-reranking framework with instruction-tuned LLMs.
The evaluation tasks are binary/multi-class classification, segmentation, and translation across five languages: Arabic, English, Finnish, Japanese, and Spanish.
Our results and analysis show the following insights:
\begin{itemize}
    \item the output probability of generating the ground-truth output is indeed good proxy for the utility of a demonstration, especially when the probability values are distributed across the whole value range with short outputs (e.g., classification),
    \item the downstream metric reward with nuanced values, on the other hand, is more robust for longer outputs (e.g., text generation tasks),
    \item our proposed incremental utility further improves ICL given contrastive training examples to effectively train the reranking model, and
    \item constrained retrieval is helpful when the retrieved candidates are imbalanced for training.
\end{itemize}

%% file: Sec-Pipelines.tex
\section{Inference and Training Pipelines}

This section describes our experimental design for the comparative study.
For a target task, we have a training set $\mathcal{T}=[e_1,\ldots,e_i,\ldots]$, where $e_i=(x'_i, y'_i)$ is an input-output pair; for example, $y'_i$ is a class label on a text classification task.

\subsection{Inference}

\paragraph{Demonstration retrieval and reranking}
Given a new input $x$, we use a demonstration-selection model $M(e_i, x)$ to predict utility of a training example $e_i$ as a demonstration.
$M$ is a cross-attention encoder that takes a concatenation of $e_i$ and $x$ as an input and then outputs a utility score.
We select demonstrations with the $k$ highest scores.
To reduce the search space for the cross-attention encoder, we use an off-the-shelf text retriever to retrieve top-$n$ training examples, according to similarity between $x$ and $x'_i$.
This is motivated by the success of the retrieval-based demonstration~\citep{icl_retrieval} and the standard retrieve-then-rerank framework~\citep{rank-t5}.
Hence we call $M$ a {\it reranking} model.

\paragraph{ICL: prompting LLMs}
We create a prompt: $\mathrm{prompt}(x, \mathcal{E}_x)$, where $\mathcal{E}_x$ is the set of the top-$k$ demonstrations selected from the $n$ retrieved examples.
The prompt is then fed into an LLM to generate a prediction as a string:
\begin{equation}
    y=\mathrm{LLM}(\mathrm{prompt}(x, \mathcal{E}_x)),
\end{equation}
where the prediction $y$ is used for task-specific evaluation.

\subsection{Training}
\label{subsec:training}

We use a utility function $u(e_i,x)$ to train the reranking model $M$; the larger the value is, the more $e_i$ is expected to contribute to ICL for $x$.
To create the training examples for $M$, we simulate the inference process within the training set; for each $x'_j\in\mathcal{T}$, we retrieve the top-$n$ set and compute $u(e,x'_j)$ with $e\in\mathcal{E}_{x'_j}$.
We ensure that $e_j$ is not in $\mathcal{E}_{x'_j}$, and consequently we have $n|\mathcal{T}|$ values of $u(e,x)$ in total.
Assuming that the value range of the utility and the reranking score is $[0.0, 1.0]$, we use logistic regression for pointwise training~\citep{bert-rerank-1,bert-rerank-2}:
\begin{eqnarray}
\label{eq:loss}
\begin{split}
    &- u(e_i,x)\log M(e_i, x) \\
    &- (1-u(e_i,x))\log (1-M(e_i, x)).
\end{split}
\end{eqnarray}

%% file: Sec-Utility.tex
\section{Utility functions}
\label{sec:inc_util}

We define multiple utility functions to investigate the effects on target tasks.
Note that the utility functions are used only in the training phase.

\input{TAB-inc_util_examples}

\input{TAB-datasets}

\subsection{Direct Utility}
\label{subsec:direct_util}

To compute $u(e,x'_j)$, it is considered to be the most straightforward to inspect how the LLM behaves given $\mathrm{prompt}(x'_j,\{e\})$.
We describe two utility functions that directly use the {\it 1-shot} result.

\paragraph{Output Probability (OP)}
We follow a widely-used approach of using the likelihood of predicting the ground-truth output~\citep{icl_retrieval_fine_2}: 
\begin{equation}
    u_{\mathrm{OP}}(e,x'_j)=p(y'_j | \mathrm{prompt}(x'_j,\{e\})),
\end{equation}
assuming that we have access to the probability by feeding the prompt and $y'_j$ together into the LLM.\footnote{\citet{icl_retrieval_fine_2} normalized the output probability values across possible output candidates for classification and multi-choice tasks. We do not apply the normalization, because the computational cost is not negligible when handling a large number of classes.}

\paragraph{Downstream Metric (DM)}
The primary goal is to improve downstream metrics of the target tasks.
As in reward optimization~\citep{seq2seq-rl}, we define the following function:
\begin{equation}
    u_{\mathrm{DM}}(e,x'_j) = R(y^*,y'_j),
\end{equation}
where $y^*=\mathrm{LLM}(\mathrm{prompt}(x'_j,\{e\}))$ is the LLM's prediction, and $R$ is a pre-defined reward function that is correlated with a task-specific metric; Section~\ref{subsec:datasets} describes a reward function for each dataset.

\subsection{Incremental Utility}
\label{subsec:inc_util}

Another way of interpreting the value of $u(e,x'_j)$ is how much {\it incremental} knowledge $e$ brings into the LLM.
For this, we would like to inspect the LLM's capability of handling $x'_j$ in the {\it 0-shot} inference:
\begin{equation}
    \mathrm{LLM}(\mathrm{prompt}(x'_j,\{\})),
\end{equation}
where $\{\}$ represents an empty demonstration set.
We can use the above-defined utility functions to evaluate the inference result; we use $u_0(x'_j)$ to denote a utility function without $e$.

For example, let us think about two cases where the use of $e$ improves the value by $0.1$: $(u_0(x'_j), u(e,x'_j))=(0.9, 1.0)$ and $(0.0, 0.1)$.
If we use $u(e,x'_j)$ alone, the former case has much larger value; however, this could underestimate the latter case's value.
We interpret those cases from different angles:
\begin{itemize}
\item[(A)] both have the same value in the absolute difference ($u(e,x'_j)-u_0(x'_j)=0.1$), or
\item[(B)] the absolute difference is only 10\% of $u(e,x'_j)$ in the former case and 100\% in the latter case.
\end{itemize}
The idea A has been explored in previous work~\citep{info-score}, while the idea B is expected to be more reasonable with the example cases.
One potential caveat is that the ratio-based approach overestimates the value, for example, with $(u_0(x'_j), u(e,x'_j))=(0.0, 0.0001)$.

We propose incremental utility $r$ to take into account all the aspects discussed above:
\begin{equation}
    \label{eq:incremental_utility}
    r(e,x'_j) = \frac{u(e,x'_j)-u_0(x'_j)}{\max(u(e,x'_j), u_0(x'_j))^\ell},
\end{equation}
where $\ell\in[0.0, 1.0]$ is a hyper-parameter.
Here are the following key features we can read:
\begin{itemize}
    \item $\ell=0.0$ corresponds to the idea A, $\ell=1.0$ to B, and the others take a balance between A and B; examples (a), (b), and (c) in Table~\ref{tab:inc_util_examples} describe the above-mentioned cases.
    \item Examples (d) and (e) in Table~\ref{tab:inc_util_examples} show the {\it symmetricity} of the function.
    \item The value range of $r$ is $[-1.0,1.0]$, and $r<0$ means that $e$ {\it negatively} affects ICL.
\end{itemize}

Finally, we linearly transform $r$ into $[0.0, 1.0]$ to define the following incremental utility function:
\begin{equation}
    \label{eq:linear_transform}
    u^{+}(e,x'_j)=\frac{r(e,x'_j)+1}{2}.
\end{equation}
This can be applicable to both the direct utility functions: $u^{+}_\mathrm{OP}$ and $u^{+}_\mathrm{DM}$.

%% file: TAB-inc_util_examples.tex
\begin{table*}[t]
\resizebox{\textwidth}{!}{
\centering
\begin{tabular}{c|cc|c|c||cccc}
\toprule
    & $u_0(x'_j)$ & $u(e,x'_j)$ & $u(e,x'_j)-u_0(x'_j)$ & $\max(u(e,x'_j), u_0(x'_j))$ & $\ell=0.0$ & $\ell=0.5$ & $\ell=0.8$ & $\ell=1.0$ \\ \hline
(a) & 0.0         & 0.1         & 0.1                   & 0.1                          & 0.1 & 0.316 & 0.631 & 1.0 \\ \hdashline
(b) & 0.9         & 1.0         & 0.1                   & 1.0                          & 0.1 & 0.1~~~~ & 0.1~~~~ & 0.1 \\ \hdashline
(c) & 0.0         & ~~~~~0.0001 & ~~~~~0.0001           & ~~~~~0.0001                  & ~~~~~0.0001 & 0.01~~ & 0.158 & 1.0  \\ \hline
(d) & 0.3         & 0.5         & 0.2                   & 0.5                          & 0.2 & 0.283 & 0.348 & 0.4 \\ \hdashline
(e) & 0.5         & 0.3         & $-$0.2~~~             & 0.5                          & $-$0.2~~~ & $-$0.283~~~ & $-$0.348~~~ & $-$0.4~~~ \\
\bottomrule 
\end{tabular}
}
\caption{Synthetic examples of $r(e,x'_j)$ in Equation~(\ref{eq:incremental_utility}) to illustrate how the incremental utility works.}
\label{tab:inc_util_examples}
\end{table*}

%% file: TAB-datasets.tex
\begin{table}[t]
\centering
\small{
\begin{tabular}{l|r|r|r}
\toprule
             & \multicolumn{1}{c|}{Train} & \multicolumn{1}{c|}{Validation} & \multicolumn{1}{c}{Test} \\ \hline 
ISD (en)     & 3,122  & 346   & 1,400 \\ \hdashline
ISD (ar)     & 2,792  & 310   & 1,400 \\ \hline
EDOS-A (en)  & 14,000 & 2,000 & 4,000 \\ \hdashline
EDOS-B (en)  & 3,398  & 486   & 970 \\ \hline
CLINC (en)   & 15,100 & 3,100 & 5,500 \\ \hline
SSENT (en)   & 1,744  & 249   & 499 \\ \hdashline
SSENT (es)   & 1,438  & 206   & 410  \\ \hline
XML-MT (ja)  & 100,033& 500   & 1,500 \\ \hdashline
XML-MT (fi)  & 97,893 & 500   & 1,500 \\
\bottomrule 
\end{tabular}
}
\caption{Dataset statistics.}
\label{tab:datasets}
\end{table}

%% file: Sec-Experiments.tex
\section{Experimental Settings}

This section describes our experimental settings; more comprehensive descriptions are in Appendix.

\subsection{LLM and Prompt}

We use Flan-PaLM 2 (L)~\citep{palm2} as our LLM.
This is an instruction-tuned model, and we follow \citet{amb_icl} to design the prompt $\mathrm{prompt}(x, \mathcal{E}_x)$ that concatenates a task instruction, demonstrations $\mathcal{E}_x$, and an input $x$.

\subsection{Tasks and Reward Functions}
\label{subsec:datasets}

We focus on NLP datasets that are {\it not} used in the Flan instruction tuning~\citep{flan-t5-paper}, across different tasks and languages; Arabic (ar), English (en), Finnish (fi), Japanese (ja), and Spanish (es) are covered.
Table~\ref{tab:datasets} shows the dataset statistics, and Table~\ref{tab:task_examples} in Appendix shows some examples.

\paragraph{Binary classification}
The task is to output a single class label for binary detection, and the value of the reward function $R(y^{*},y'_j)$ is $1.0$ if $y^{*}=y'_j$, and $0.0$ otherwise.
The evaluation metric is corpus-level detection F1.
\begin{itemize}
    \item \textbf{ISD} is a dataset for detection of ``sarcasm'' text in English and Arabic~\citep{isd}.
    \item \textbf{EDOS-A} is a dataset for detection of ``sexist'' text in Egnlish~\citep{edos}.
\end{itemize}

\paragraph{Multi-class classification}
The task is to output a single class label for fine-grained classification, and $R(y^{*},y'_j)$ is the same as that of binary classification.
\begin{itemize}
    \item \textbf{EDOS-B} is a dataset for 4-way fine-grained classification about the sexist text in English~\citep{edos}; the evaluation metric is the macro F1.
    \item \textbf{CLINC} is a dataset for 150-way intent classification of user input text in English, which also tests out-of-domain detection~\citep{clinc}; the evaluation metric is the joint accuracy proposed in \citet{dnnc}.
\end{itemize}

\paragraph{Segmentation}
The task is to output a tagged version of an input text.
$R(y^{*},y'_j)$ is a word-level F1, and the evaluation metric is the same.
\begin{itemize}
    \item \textbf{SSENT} is a decomposed subtask of structured sentiment analysis~\citep{ssent}, and we use the OpeNER portion~\citep{opener} in English and Spanish.
\end{itemize}

\paragraph{Translation}
The task is to translate text from a language to another, and $R(y^{*},y'_j)$ is example-level GLEU~\citep{gnmt}.
The evaluation metric is BLEU~\citep{bleu}.
\begin{itemize}
    \item \textbf{XML-MT} is a dataset for translation of XML-tagged Web documentation~\citep{xml-mt}, and we take the English-to-Japanese/Finnish subsets.
\end{itemize}

\subsection{Demonstration Retrieval and Reranking}

As the off-the-shelf text retriever, we use a generic t5x retriever~\citep{t5x_retrieval} used in previous work~\citep{aditi_qgen,amb_icl}, which has a multilingual capability based on mT5~\citep{mt5paper}.
We also use mT5 (following RankT5~\citep{rank-t5}) to train the reranking model, which can be seamlessly applied to the different languages' data.

We use $n=10$ to train and validate the reranking model, and use $n=50$ for the final test evaluation to investigate the generalization ability.
We have run training of the reranking model in about 100 configurations in total, for different datasets, hyperparameter search of $\ell$ in Equation~(\ref{eq:incremental_utility}), and checkpoint selection with evaluation on the validation sets, before touching the test sets.

\subsection{Methods to be Compared}

For the evaluation, we report results by the following methods including the standard baselines and our reranking methods:
\begin{itemize}
    \item ``\textbf{0-shot}'' is a baseline to know how the LLM performs only with the task instruction.
    \item ``\textbf{RETR}'' is another baseline to know how the LLM performs by simply selecting the top-$k$ retrieved demonstrations.
    \item ``$u_{\mathrm{OP}}$'' and ``$u_{\mathrm{DM}}$'' are our main baselines by the reranking models trained with the direct utility functions (Section~\ref{subsec:direct_util}).
    \item ``$u_{\mathrm{OP}}^+$'' and ``$u_{\mathrm{DM}}^+$'' are the main methods for our analysis, by the reranking models trained with the incremental utility functions (Section~\ref{subsec:inc_util}).
\end{itemize}

When constructing the prompts, we order the demonstrations according to the retrieval scores for RETR, and according to the reranking scores for the reranking-based methods.

\input{TAB-main_results}

\section{Results and Discussions}
\label{sec:results}

Table~\ref{tab:main_table} shows the $k$-shot ICL results on the test sets, where we use $k=1,3,5$ for all the datasets, except for XML-MT with $k=1,2,3$ due to its prompt being too long.
All the evaluation scores range in [0, 100].
We also conduct a cross-lingual experiment with SSENT; we use the English training examples for retrieval, and directly apply the English reranking models to the Spanish test set.

\begin{figure}[t]
\includegraphics[width=\linewidth]{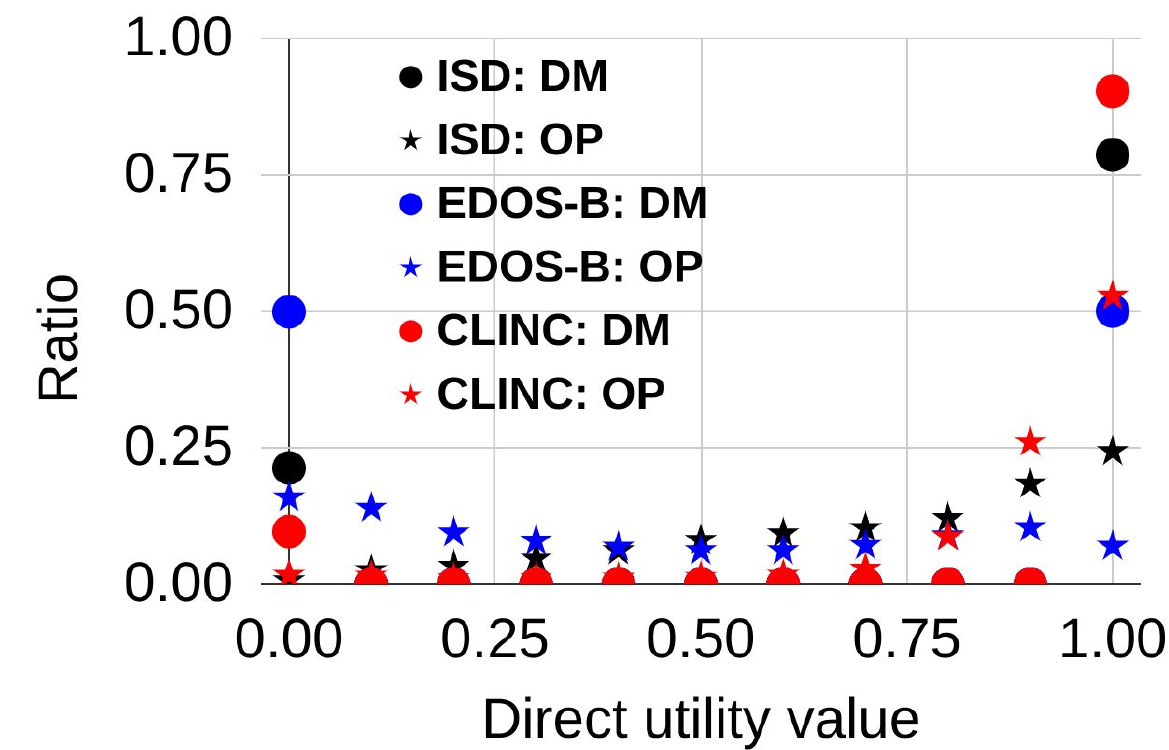}
\caption{Distributions of $u_\mathrm{DM}$ and $u_\mathrm{OP}$ on the training sets of the three classification datasets.}
\label{fig:ditect_utility_dist_classification}
\end{figure}

\begin{figure}[t]
\includegraphics[width=\linewidth]{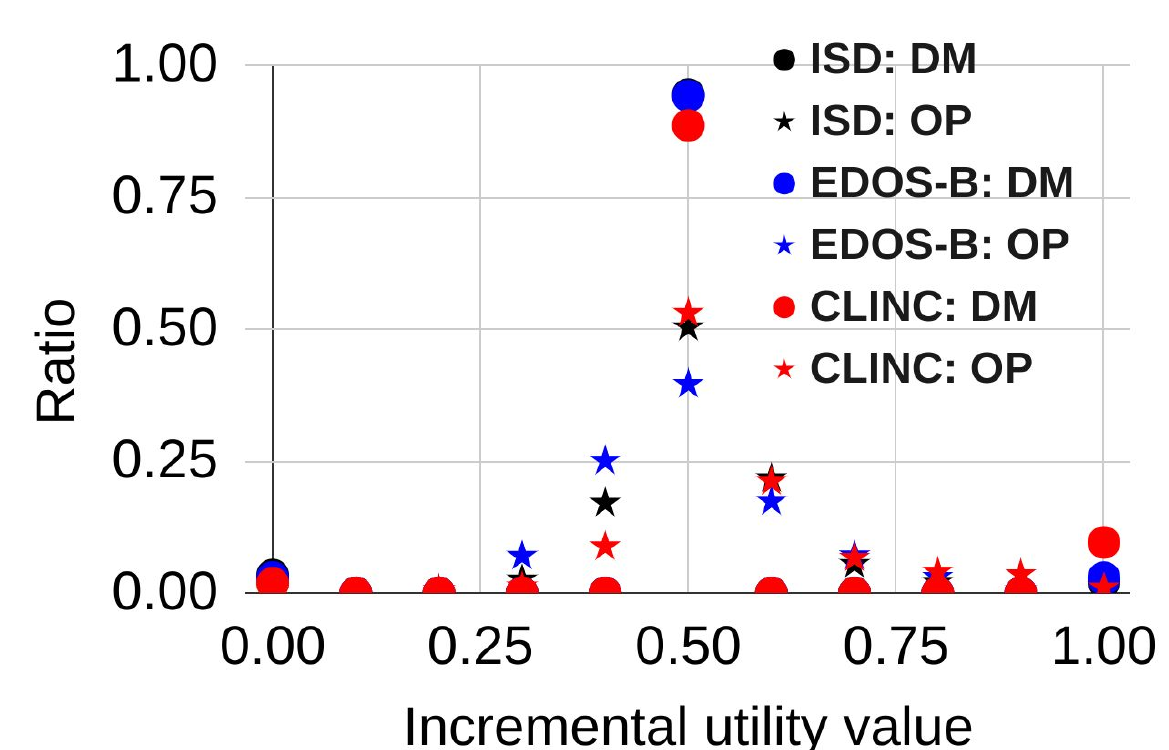}
\caption{Distributions of $u_\mathrm{DM}^+$ and $u_\mathrm{OP}^+$ on the training sets of the three classification datasets.}
\label{fig:incremental_utility_dist_classification}
\end{figure}

We first summarize general observations, and then show our analysis to provide insights.

\begin{itemize}
\item RETR ICL outperforms the 0-shot baseline on all the datasets, except for ISD (en). This indicates that the general input-text similarity would not always find useful demonstrations.
\item Reranking (with at least one utility function) improves upon RETR. Only one exception is ISD (ar) with $k=1$, but $k=3$ and $k=5$ still show the benefit.
\item The OP utility tends to be more effective than the DM utility, especially on the classification tasks, while we can see potential advantages of DM on SSENT (es) and XML-MT.
\item Our proposed incremental utility tends to improve upon the direct utility, especially with the OP utility.
\item English-based reranking is well transferred to Spanish on SSENT. This is encouraging, because it is often the case that we only have English resources for training.
\item As motivated and expected in Section~\ref{subsec:inc_util}, the incremental utility functions work the best with $0.0<\ell<1.0$, especially $\ell=0.8$ (more details in Appendix~\ref{app:inc_util}).
\end{itemize}

\subsection{Analysis on Classification Tasks}
\label{subsec:analysis_classification}

A clear advantage of OP on the classification task is to provide denser training signals than DM.
Figure~\ref{fig:ditect_utility_dist_classification} (best viewed in color) shows distributions of $u_\mathrm{DM}$ and $u_\mathrm{OP}$ on the training sets of ISD (en), EDOS-B, and CLINC; the OP-based utility values are well distributed across the value range.
Figure~\ref{fig:incremental_utility_dist_classification} then shows distributions of $u_\mathrm{DM}^+$ and $u_\mathrm{OP}^+$, and we can see the consistent trend.

\paragraph{Existence of negative incremental values}
A notable observation is that not all the demonstrations bring positive incremental values, as shown in Figure~\ref{fig:incremental_utility_dist_classification}.
Based on the definition of the incremental utility function (Equation~(\ref{eq:linear_transform})), a demonstration has a {\it negative} effect when its incremental utility value is less than 0.5.
It is interesting to see that even the top-10 retrieved candidates can have negative impacts in ICL.

\paragraph{When is incremental utility effective?}
Now that we have seen the difference of the distributions of the OP and DM values, a remaining question is why the incremental utility is not always effective.
Specifically, Table~\ref{tab:main_table} shows that $u_\mathrm{OP}^+$ performs worse than $u_\mathrm{OP}$ on ISD (ar) and CLINC.
To explain the results, we count the number of {\it contrastive} training examples that meet the following criterion:
\begin{itemize}
    \item[] a training example $e$ is considered to be contrastive, if its corresponding $n$ ($=10$) demonstration candidates cover both the $u(e,x'_j)>u_0(x'_j)$ and $u(e,x'_j)<u_0(x'_j)$ cases.
\end{itemize}
We expect such a training example to contribute to effective contrast of the demonstrations.
Table~\ref{tab:effective_analysis_classification} shows the counting results, along with the improvement (averaged across $k=1,3,5$) over $u_\mathrm{OP}$ by $u_\mathrm{OP}^+$ in Table~\ref{tab:main_table}, and we can see a positive correlation as expected.

\begin{table}[t]
\small{
\centering
\begin{tabular}{l|c|c}
\toprule
& Contrastive examples & Improvement \\
&                      & by $u_\mathrm{OP}^+$ \\ \hline
EDOS-B   & 88.6\% & +8.79 \\ \hline
ISD (en) & 62.8\% & +0.18 \\ \hline
EDOS-A   & 56.2\% & +0.71 \\ \hline
CLINC    & 49.1\% & -0.71 \\ \hline
ISD (ar) & 37.9\% & -0.82 \\
\bottomrule 
\end{tabular}

}
\caption{Importance of the contrastive examples (with OP) for the classification tasks.}
\label{tab:effective_analysis_classification}
\end{table}

\begin{table}[t]
\small{
\centering
\begin{tabular}{l|c|c|c}
\toprule
                      & $n'=10$ & $n'=30$ & $n'=50$            \\ \hline
Contrastive examples  & 39.5\%  & 49.3\% & 53.5\%   \\
\bottomrule 
\end{tabular}

}
\caption{The effects of $n'$ on ISD (ar).}
\label{tab:comparison_n}
\vspace{4mm}
\small{
\centering
\begin{tabular}{l|c|c}
\toprule
                   & Original            & Constrained            \\ \hline
$u_\mathrm{OP}$    & 52.92, 57.14, 57.53 & 53.31, \textbf{58.25}, \textbf{59.17}    \\ \hdashline
$u_\mathrm{OP}^+$  & 52.44, 55.87, 56.82 & \textbf{54.00}, 57.29, 57.10   \\
\bottomrule 
\end{tabular}

}
\caption{Comparison between original (Table~\ref{tab:main_table}) and constrained retrieval strategies on ISD (ar).}
\label{tab:comparison_retrieval}
\end{table}

\paragraph{Can we increase contrastive examples?}
We then discuss how to increase the number of the contrastive examples.
Inspired by constrained retrieval~\citep{amb_icl}, we modify our simple retrieval process with the following steps:
\begin{itemize}
    \item[1.] retrieving top-$N$ ($N>n$) demonstration candidates for a training example $e$,
    \item[2.] selecting $n'$ candidates according to the retrieval scores, by ensuring that all the class labels' examples are equally selected,
    \item[3.] computing the direct utility scores of all the $n'$ candidates, and
    \item[4.] selecting $n$ (=10) candidates to make $e$ a contrastive example if possible.
\end{itemize}
The step 4. is optional to control the number of candidates to be used for training the reranking model.
For ISD (ar), we set $N=300$\footnote{$N$ is a dataset-dependent (or retriever-dependent) hyperparameter, and we suggest to set the value to maximize the class label coverage of the retrieved candidates.} and $n'=50$.
Table~\ref{tab:comparison_n} shows how the number of the contrastive examples increases as expected, and Table~\ref{tab:comparison_retrieval} shows the ISD (ar) test evaluation results.
It is notable that the constrained retrieval lets $u^+_\mathrm{OP}$ perform the best with $k=1$.

\paragraph{Necessity of compositional utility}
We see in Table~\ref{tab:comparison_retrieval} that $u_\mathrm{OP}$ still performs better than $u^+_\mathrm{OP}$ with $k=3,5$; however, this is less controllable in our work because the utility values are defined for each demonstration independently.
Therefore, it is a crucial next step to investigate compositional effects of adding multiple demonstrations~\citep{coverage-icl,compo-icl}.

\paragraph{General instruction}
In summary, we provide the following instruction for classification tasks:
\begin{itemize}
    \item using $u_\mathrm{OP}$ if the number of the contrastive training examples is low (e.g., less than 50\%), and otherwise $u_\mathrm{OP}^+$, and
    \item using the constrained retrieval if there are much less contrastive training examples (e.g., less than 40\%).
\end{itemize}

\subsection{Analysis on Non-Classification Tasks}

\begin{figure}[t]
\includegraphics[width=\linewidth]{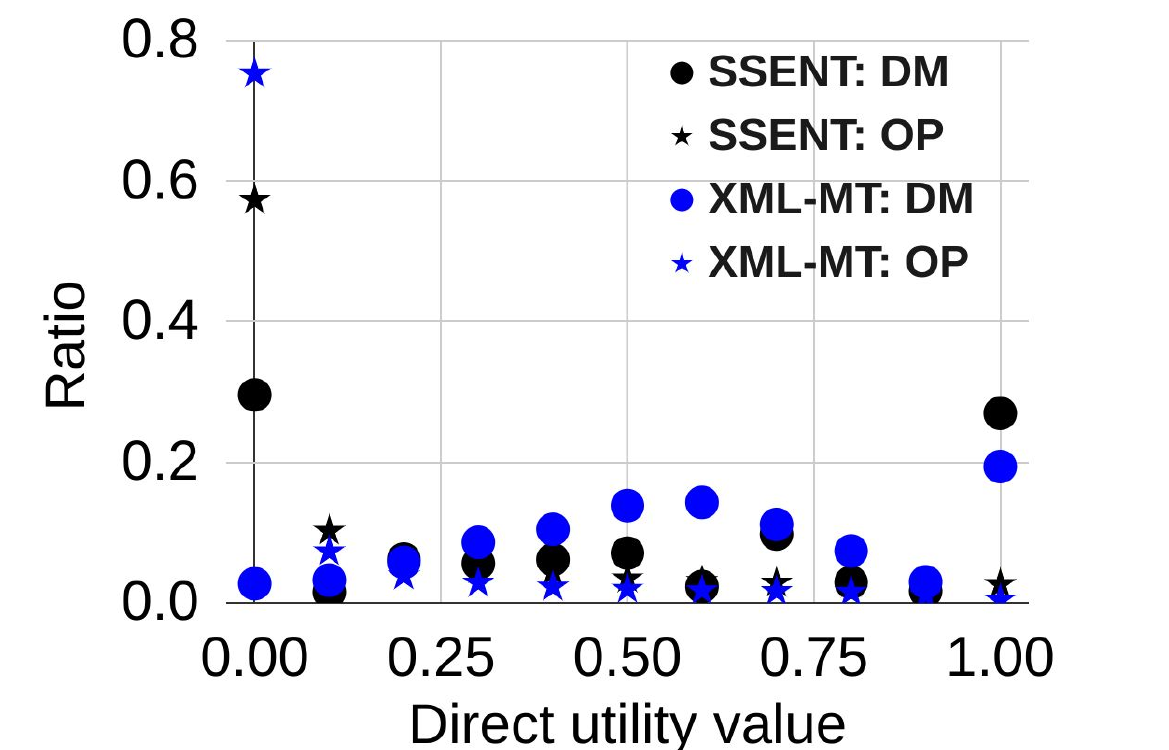}
\caption{Distributions of $u_\mathrm{DM}$ and $u_\mathrm{OP}$ on the training sets of the two non-classification datasets.}
\label{fig:ditect_utility_dist_generation}
\end{figure}

\begin{figure}[t]
\includegraphics[width=\linewidth]{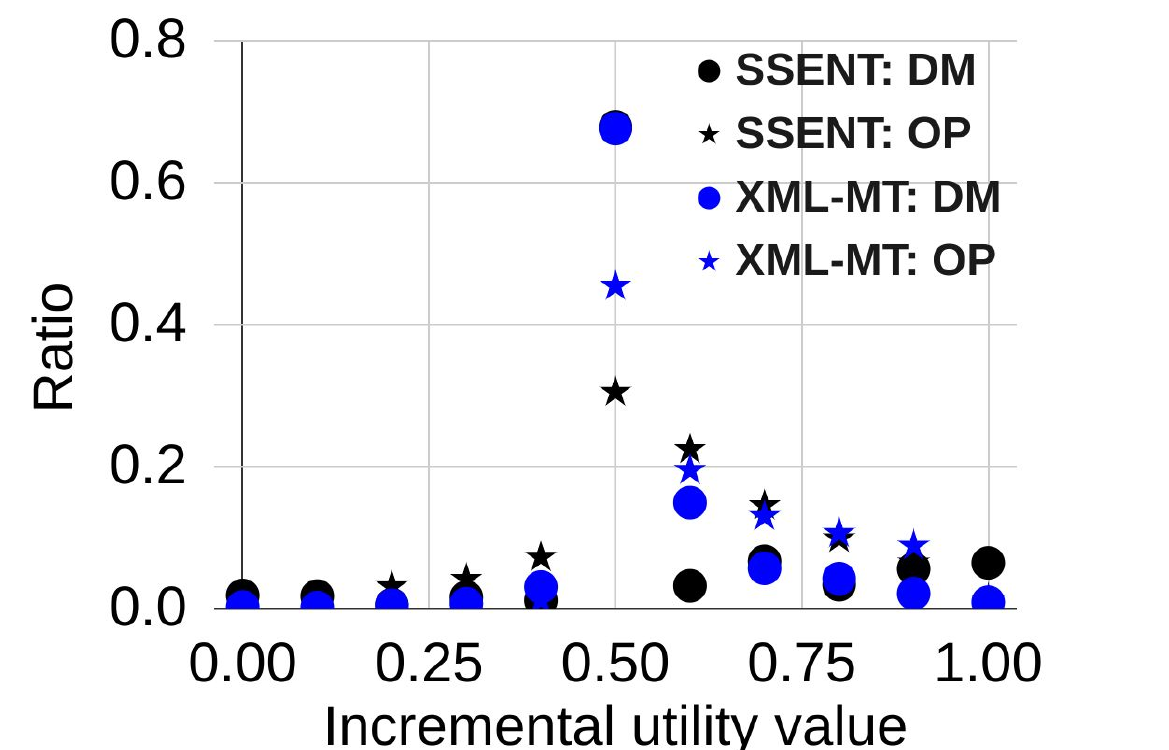}
\caption{Distributions of $u_\mathrm{DM}^+$ and $u_\mathrm{OP}^+$ on the training sets of the two non-classification datasets.}
\label{fig:incremental_utility_dist_generation}
\end{figure}

Figure~\ref{fig:ditect_utility_dist_generation} and Figure~\ref{fig:incremental_utility_dist_generation} (best viewed in color) show the distributions of the utility values on SSENT (es) and XML-MT (ja) as in Section~\ref{subsec:analysis_classification}.
Unlike the classification tasks, the direct utility values are well distributed by the task-oriented reward.

\begin{table}[t]
\small{
\centering
\begin{tabular}{l|c|c}
\toprule
& Contrastive examples & Improvement \\
&                      & by $u_\mathrm{OP}^+$ \\ \hline
SSENT (en)   & 56.4\% & +1.76 \\ \hline
SSENT (es)   & 56.4\% & +1.08 \\ \hline
XML-MT (fi)  & 44.0\% & +1.09 \\ \hline
XML-MT (ja)  & 30.2\% & -0.03 \\
\bottomrule 
\end{tabular}

}
\caption{Importance of the contrastive examples (with OP) for the non-classification tasks.}
\label{tab:effective_analysis_generation}
\vspace{4mm}
\small{
\centering
\begin{tabular}{l|c|c}
\toprule
& Contrastive examples & Improvement \\
&                      & $u_\mathrm{DM}^+$ \\ \hline
XML-MT (ja)  & 28.4\% & -0.11 \\ \hline
XML-MT (fi)  & 27.1\% & +0.14 \\ \hline
SSENT (es)   & 0.06\% & -1.01  \\ \hline
SSENT (en)   & 0.05\% & -0.13 \\
\bottomrule 
\end{tabular}

}
\caption{Importance of the contrastive examples (with DM) for the non-classification tasks.}
\label{tab:effective_analysis_generation_dm}
\end{table}

\paragraph{Is incremental utility effective with OP?}
One observation is that 60-80\% of the $u_\mathrm{OP}$ values fall into the [0.0, 0.05) bucket; this is presumably because the LLM is not finetuned with the task formats of SSENT and XML-MT, and generating long texts leads to lower probability values in general~\citep{gnmt}.\footnote{Even if we would be able to re-scale the output probability values by the output sequence lengths, it is not always possible to have access to token-level information of the outputs. For example, some of the recent LLMs only provide APIs.}
This would hinder the effectiveness of $u_\mathrm{OP}$ on the non-classification tasks, and actually Table~\ref{tab:main_table} shows that $u_\mathrm{DM}$ consistently outperforms $u_\mathrm{OP}$ (with $k=1,2,3$) on XML-MT.
However, as in the classification tasks, we can see that $u^+_\mathrm{OP}$ helps improve the scores, and Table~\ref{tab:effective_analysis_generation} shows the effects of the contrastive examples.

\paragraph{Limitation of DM}
We have seen the potential advantage of DM by the comparison between $u_\mathrm{DM}$ and $u_\mathrm{OP}$ on the SSENT and XML-MT datasets.
This is an expected observation that $u_\mathrm{DM}$ works well on the non-classification tasks based on the nuanced values as the reward, only by having access to the output strings.
However, $u_\mathrm{DM}^+$ does not improve the scores in most of the cases (shown in Table~\ref{tab:main_table}).
Table~\ref{tab:effective_analysis_generation_dm} shows the effects of the contrastive training examples from a viewpoint of DM.
Surprisingly, there are much less contrastive examples than those with OP.

\paragraph{Necessity of compositional utility}
One reason why we could not effectively create the contrastive training examples is that, the LLM's actual prediction is not easily affected by only one demonstration (i.e., $k=1$).
For example, if we inspect the LLM's predictions on the validation set of SSENT (en), only 27.4\% of the independent demonstrations lead to different predictions than those with the 0-shot setting.
If we include demonstrations that can modify the LLM's predictions with $k=2$, the ratio increases to 40.1\%.
We have then missed potentially useful signals for the utility estimation.
This indicates that it is necessary for future work to estimate the utility of the demonstrations with $k>1$ as also discussed in Section~\ref{subsec:analysis_classification}.

\paragraph{General instruction}

In summary, we provide the following instruction for non-classification tasks:

\begin{itemize}
\item using $u_\mathrm{DM}$ if well-defined nuanced metrics for the reward can be implemented, and
\item considering the use of $u^+_\mathrm{OP}$ otherwise, as in the classification tasks.
\end{itemize}

\subsection{Demonstrations with High Utility}

We have investigated the effectiveness of incremental utility, and in this section, we focus on how the utility values are related to the demonstrations' ground-truth outputs.
For the sake of simplicity, we take the classification datasets in this analysis.

For each dataset, we list all the $n|\mathcal{T}|$ input-demonstration pairs from the training set as in Section~\ref{subsec:training}.
We then sort the pairs to select the top-$m$\% of them either by $u_\mathrm{OP}$ or $u^+_\mathrm{OP}$, and calculate the ratio of the pairs whose demonstration labels match with their paired input's labels.

Table~\ref{tab:high_util_demos} shows the results, and overall, we can see that $u_\mathrm{OP}$ and $u^+_\mathrm{OP}$ rank the pairs differently as expected in Section~\ref{subsec:inc_util}.
Judging from Table~\ref{tab:main_table} and Table~\ref{tab:effective_analysis_classification}, $u_\mathrm{OP}^+$ performed the best on EDOS-B, and we can see much more matched labels when sorted by $u_\mathrm{OP}^+$ than $u_\mathrm{OP}$.
This is considered to be a factor of the improvement, related to previous studies about the importance of the demonstrations' labels~\citep{icl-bias-analysis-min,amb_icl}, while there are other factors like coverage~\citep{coverage-icl} or compositionality~\citep{compo-icl}.

\begin{table}[t]
\small{
\centering
\begin{tabular}{ll|c|c|c}
\toprule
    &  & Top-10\%     & Top-30\%     & Top-50\%           \\ \hline
ISD (en) & $u_\mathrm{OP}$      & 68.7\% & 67.1\% & 64.8\%   \\ \hdashline
         & $u_\mathrm{OP}^+$    & 68.7\% & 67.2\% & 65.1\%    \\ \hline
ISD (ar) & $u_\mathrm{OP}$      & 93.6\% & 89.4\% & 86.3\%   \\ \hdashline
         & $u_\mathrm{OP}^+$    & 81.1\% & 81.2\% & 81.7\%    \\ \hline
EDOS-A   & $u_\mathrm{OP}$      & 69.3\% & 66.7\% & 66.3\%   \\ \hdashline
         & $u_\mathrm{OP}^+$    & 72.5\% & 74.1\% & 73.7\%    \\ \hline
EDOS-B   & $u_\mathrm{OP}$      & 57.4\% & 53.8\% & 51.1\%   \\ \hdashline
         & $u_\mathrm{OP}^+$    & 93.3\% & 72.3\% & 62.8\%    \\ \hline
CLINC    & $u_\mathrm{OP}$      & 86.2\% & 86.9\% & 86.1\%   \\ \hdashline
         & $u_\mathrm{OP}^+$    & 85.3\% & 78.4\% & 78.7\%    \\
\bottomrule 
\end{tabular}

}
\caption{Label matching ratio with high utility.}
\label{tab:high_util_demos}
\end{table}

\subsection{Generalization Ability}
\label{subsec:cross_lm}

We have used Flan-PaLM 2 (L) as an LLM, and the t5x retriever as a baseline retriever; these are the core components in our evaluation pipeline.
The last question we would like to answer here is whether our empirical findings are generalized by replacing the LLM or the retriever.

\subsubsection{LLMs}
We conduct experiments by simply replacing Flan-PaLM 2 (L) with another instruction-tuned model, Flan-T5 XXL~\citep{flan-t5-paper}, or a {\it non}-instruction-tuned model, GPT-J~\citep{gpt-j}.
For Flan-T5, we use exactly the same prompt designs, and for GPT-J, we use different prompt designs as detailed in Appendix~\ref{app:gptj}.

Table~\ref{tab:flan_t5} reports the results on EDOS-B, and we can see the consistent advantage of $u^{+}_\mathrm{OP}$ on the dataset.
One interesting observation is that GPT-J performs much worse than the instruction-tuned models in the baseline results, but the scores are dramatically improved by the reranked demonstrations.
The results show that different LLMs make use of the demonstrations differently, which would be dependent on the target tasks, inherent biases of the LLMs, and the pre-training strategies.
For example, the ``Animosity'' class on the dataset is much better predicted by GPT-J than the others.
As a result, from a viewpoint of the macro F1 metric, GPT-J has an advantage, because performing poorly on a class significantly hurts the metric.

\begin{table}[t]
\small{
\centering
\begin{tabular}{l|c|c}

\toprule
                   & Flan-T5 XXL         & GPT-J            \\ \hline
0-shot             & 35.28               & 19.78               \\ \hline
RETR               & 36.59, 38.67, 38.73 & 37.12, 35.16, 29.68    \\ \hline
$u_\mathrm{OP}$    & 36.02, 38.81, 39.71 & 55.78, 57.23, 56.40    \\ \hdashline
$u_\mathrm{OP}^+$  & \textbf{41.30}, \textbf{45.42}, \textbf{46.89} & \textbf{58.97}, \textbf{59.15}, \textbf{58.54}   \\ \hline
$u_\mathrm{DM}$    & 38.10, 40.13, 41.98 & 51.28, 53.08, 53.60   \\ \hdashline
$u_\mathrm{DM}^+$  & 36.85, 39.89, 40.70 & 52.33, 54.25, 54.55   \\
\bottomrule

\end{tabular}

}
\caption{Results on EDOS-B by using Flan-T5 XXL or GPT-J as the LLM component.}
\label{tab:flan_t5}
\end{table}

\begin{table}[t]
\small{
\centering
\begin{tabular}{l|c|c}

\toprule
                   & EDOS-B (en)         & SSENT (en)            \\ \hline
0-shot             & 41.49               & 34.06               \\ \hline
RETR               & 40.81, 46.05, 49.47 & 45.20, 50.75, 54.43    \\ \hline
$u_\mathrm{OP}$    & 40.81, 43.57, 46.41 & 49.79, 54.75, 58.50    \\ \hdashline
$u_\mathrm{OP}^+$  & \textbf{44.70}, \textbf{51.01}, \textbf{53.31} & \textbf{51.60}, \textbf{56.96}, \textbf{60.21}   \\ \hline
$u_\mathrm{DM}$    & 40.71, 44.00, 43.30 & 49.19, 54.15, 57.13   \\ \hdashline
$u_\mathrm{DM}^+$  & 42.04, 43.75, 46.18 & 48.14, 54.39, 57.95   \\
\bottomrule

\end{tabular}

}
\caption{Results on EDOS-B and SSENT (en) by replacing the t5x retriever with a TF-IDF retriever only at the test time.}
\label{tab:tfidf}
\end{table}

\subsubsection{Retrievers}

We then conduct another set of the experiments by replacing the t5x retriever with a TF-IDF retriever\footnote{\url{https://scikit-learn.org/stable/modules/generated/sklearn.feature_extraction.text.TfidfVectorizer.html}} at the test time, to see how the reranking model (trained with the t5x retriever) works with different candidate sets.
Table~\ref{tab:tfidf} shows the results on EDOS-B and SSENT (en).
We can see that the reranking model works even with the TF-IDF-based candidates, showing the robustness of the reranking with the proposed utility functions.

%% file: TAB-main_results.tex
\begin{table*}[t]
\small{
\centering
\begin{tabular}{l|c|c|c|c|c}
\toprule
                   & \multicolumn{3}{c|}{\textbf{Binary classification}} & \multicolumn{2}{c}{\textbf{Multi-class classification}} \\
                   & ISD (en)            & ISD (ar)            & EDOS-A (en)         & EDOS-B (en)         & CLINC (en)  \\ \hline
0-shot             & 58.54               &  43.43              & 57.07               & 41.49               & 86.52 \\ \hline
RETR                & 55.35, 56.98, 58.61 & \textbf{53.07}, 55.81, 55.57 & 61.78, 65.74, 67.70 & 43.10, 46.14, 48.45 & 91.89, 92.12, 92.48  \\ \hline
$u_\mathrm{OP}$    & \textbf{58.38}, 59.51, 59.84 & 52.92, \textbf{57.14}, \textbf{57.53} & 64.93, 69.06, 70.77 & 42.05, 44.58, 47.56 & 93.32, \textbf{94.30}, \textbf{94.22}  \\ \hdashline
$u_\mathrm{OP}^+$  & 58.11, \textbf{59.75}, \textbf{60.41} & 52.44, 55.87, 56.82 & \textbf{65.15}, \textbf{70.17}, \textbf{71.58} & \textbf{47.85}, \textbf{55.04}, \textbf{57.66} & 92.63, 93.20, 93.89 \\ \hline
$u_\mathrm{DM}$    & 57.82, 58.97, 60.06 & 51.91, 55.61, 56.22 & 62.53, 66.07, 67.54 & 42.33, 45.48, 46.04 & \textbf{93.62}, 93.85, 93.55  \\ \hdashline
$u_\mathrm{DM}^+$  & 57.36, 59.27, 59.09 & 51.73, 55.82, 57.09 & 63.75, 67.16, 69.26 & 42.89, 48.73, 51.12 & 92.01, 92.35, 92.68  \\
\bottomrule 
\end{tabular}
\begin{tabular}{l|c|c|c|c|c}
                   & \multicolumn{3}{c|}{\textbf{Segmentation}} & \multicolumn{2}{c}{\textbf{Translation}} \\
                   & SSENT (en)          & SSENT (es)          & SSENT (en$\rightarrow$es)      & XML-MT (ja) & XML-MT (fi)    \\ \hline
0-shot             & 34.06               &  27.88              & 27.88               & 48.04                      & 33.45  \\ \hline
RETR                & 46.77, 50.60, 53.82 & 43.30, 49.71, 52.16 & 35.09, 36.59, 39.39 & 61.11, 62.80, 64.26        & 45.73, 47.50, 48.24 \\ \hline                   
$u_\mathrm{OP}$    & 49.61, 55.42, 58.74 & 44.16, 51.89, 54.49 & 35.46, 36.73, 41.13  & 63.70, 65.35, 66.32 & 46.82, 48.10, 48.70   \\ \hdashline
$u_\mathrm{OP}^+$  & \textbf{52.03}, \textbf{56.69}, \textbf{60.34} & 44.44, 53.56, \textbf{55.78} & 36.70, \textbf{39.58}, \textbf{42.27} & 63.53, 65.61, 66.14 & 47.67, \textbf{49.41}, 49.82  \\ \hline
$u_\mathrm{DM}$    & 50.76, 53.53, 57.18 & \textbf{46.92}, \textbf{53.66}, 53.88 & \textbf{37.07}, 38.00, 40.49 & \textbf{64.20}, \textbf{66.09}, 66.49 & 48.01, 49.20, 49.88   \\ \hdashline
$u_\mathrm{DM}^+$  & 49.15, 54.20, 57.74 & 45.54, 52.46, 53.43 & 35.54, 36.42, 40.14 & 64.02, 65.86, \textbf{66.56} & \textbf{48.34}, 49.26, \textbf{49.92}  \\
\bottomrule
\end{tabular}

}
\caption{Test results. Each cell reports scores with $k=1,3,5$ (except for XML-MT with $k=1,2,3$).}
\label{tab:main_table}
\end{table*}

%% file: Sec-Related.tex
\section{Related Work}

We discuss relationships between our work and previous work from several viewpoints.
Interested readers may refer to a survey by \citet{luo-survey}.

\paragraph{Utility function}
Recent work~\citep{icl_retrieval_fine_1,icl_retrieval_fine_2,icl_retrieval_fine_3,icl_retrieval_fine_4} has proposed finetuning text retrieval models by estimating direct utility of a demonstration by $u_\mathrm{OP}$-like utility functions.
While they showed the effectiveness, we have made a deeper dive into understanding the effects of the different utility functions, especially with the enhance of incremental utility.

\paragraph{Loss function}
To finetune the retrieval models, some used contrastive learning to contrast a high-utility demonstration against a number of low-utility demonstrations~\citep{icl_retrieval_fine_1,icl_retrieval_fine_3,icl_retrieval_fine_4}, and others used ranking losses~\citep{icl_retrieval_fine_2}.
Contrastive learning~\citep{dpr-paper} and ranking losses~\citep{rank-t5} are widely used for document retrieval and reranking, but simple pointwise regression has shown to be effective as well~\citep{bert-rerank-1,bert-rerank-2,asai2020learning}.
We employed the pointwise regression (Equation~(\ref{eq:loss})), based on our observations that it significantly improves ICL upon the RETR baseline.
Instead, we focused on the contrast between the 0-shot and 1-shot performances to derive the incremental utility functions.

\paragraph{Overall pipeline}
\citet{lin2023radit} have taken one step further to finetune LLMs with a retriever, while our work has focused on using an LLM as it is.
Within this framework, it is a common practice to directly finetune a dense text retrieval model~\cite{dpr-paper} for the demonstration retrieval~\citep{icl_retrieval_fine_1,icl_retrieval_fine_2,icl_retrieval_fine_3}, while \citet{icl_retrieval_fine_4} employed a two-step approach by knowledge distillation with a cross-attention reward model (similar to our reranking model).
By contrast, we followed another major framework with retrieval and reranking~\citep{rank-t5}, where we use a generic text retriever for all the different settings.
This is useful in making our experiments controllable for analysis, in that the reranking models are always applied to the consistent candidate sets.
One potential drawback of the cascaded pipeline is that the search space is limited; still, we have shown that there is large room for improvement even within the limited search space.

\paragraph{Constrained retrieval and selection}
This paper has focused on using feedback from LLMs to train the reranking models.
Another line of related work is to improve the demonstration retrieval/selection process without finetuning additional models.
\citet{amb_icl} take into account difficulty of each demonstration and entropy of the LLMs' prediction,
\citet{diverse-icl} select demonstrations based on uncertainty and diversity,
\citet{active-icl} adopt active learning, and
\citet{coverage-icl} propose to increase coverage of information about inputs.
In Section~\ref{subsec:analysis_classification}, we have shown that a simple constrained retrieval has the potential to improve the reranking models, and it is an interesting direction to incorporate those kinds of techniques into the training of the reranking models.

\paragraph{Unified modeling}
We conducted our analysis on the different tasks in several languages, and all the reranking models were trained independently.
\citet{icl_retrieval_fine_2} made an interesting attempt to finetune a retriever across different tasks, and showed promising results.
Then another interesting future direction is to integrate our findings into such a unified framework; for example, instruction-based demonstration retriever/reranker is worth investigating by following \citet{asai-etal-2023-task}.

%% file: Sec-Conclusion.tex
\section{Conclusion}
This paper has investigated how the output probability and downstream metric affect the utility estimation of demonstrations for ICL.
Our in-depth analysis has shown that the output probability is robust on the classification tasks, and the downstream metric is robust especially when the output probability is not well distributed across the whole value range.
Furthermore, we provided discussions about why and when our proposed incremental utility helps improve the ICL results.

%% file: Sec-Extra.tex
\section*{Acknowledgements}
First of all, we thank Aditi Chaudhary and Krishna Srinivasan for their contributions to the basics of the ICL work in our team.
We also appreciate discussions with Lingyu Gao about her internship project and related work.
Feedbacks from Zhuyun Dai and Tania Bedrax-Weiss were also valuable in polishing the draft.
Lastly, we would like to thank anonymous ARR reviewers for their constructive feedbacks.

\section*{Limitations}

\paragraph{Inference-time cost}
We adopted the retrieval-reranking framework, and the reranking model is a large cross-attention encoder model based on mT5 XXL.
This is known to be much slower than another standard framework with dense retrievers~\citep{dpr-paper}.
The main goal in this paper is to investigate how to transfer the feedback signals from the LLMs to the demonstration selection, and this is still an open research question.
Therefore, we first prioritized having an enough model capacity instead of restricting the inference-time cost, as in previous work on document retrieval~\citep{asai2020learning}.
Once it gets practically useful, we can consider applying knowledge distillation~\citep{icl_retrieval_fine_4}, or directly training dense retriever frameworks~\citep{icl_retrieval_fine_1}.

\section*{Ethics Statement}

\paragraph{Inherent biases}
It is known that LLMs have inherent biases that affect ICL results~\citep{icl-bias-analysis,icl-bias-analysis-min}.
We also observed that specific labels are more generated than others on the classification tasks, and the trends are different among different LLMs like PaLM 2 and GPT-J.
It is then possible that the biases are transferred to the reranking models, and potentially we can further improve the results by debiasing the feedback signals.
This has not been addressed in this paper, and is left for future work.

\paragraph{Contents of the datasets}
We used publicly available datasets that have been properly processed for privacy concerns.
The EDOS dataset contains offensive contents as described in \citet{edos}, and we intentionally avoided showing examples from the dataset.
We do not intend to enhance the offensive contents in this paper, and instead solely focus on how the LLMs work on the challenging fine-grained classification.

%% file: Sec-Appendix.tex
\appendix

\section*{Appendix}

\section{Prompt Design for Flan Models}
We use the prompt template defined in \citet{amb_icl} for the instruction-tuned LLMs in our experiments.
We then need to have a task definition for each dataset with an input-output format.

\section{Datasets}

\input{TAB-datasets_examples}

Table~\ref{tab:task_examples} shows examples of input-output pairs from the datasets used in our experiments.

\subsection{ISD}
\label{app:data_isd}

The task definition of ISD is as follows:
\newline

\noindent
{\tt The goal of this task is to identify if an input text is sarcastic or non-sarcastic.}
\newline

\noindent
Our initial analysis showed that the LLMs in our experiments have inherent bias towards generating ``sarcastic'' and ``non-sarcastic'' instead of ``Sarcastic'' and ``Non-sarcastic,'' and then we decided to use the all-lowercased label strings.
The same task definition is shared in Arabic and English.

\subsection{EDOS-A}
\label{app:data_edos_a}

The task definition of EDS-A is as follows:
\newline

\noindent
{\tt The goal of this task is to identify if an input text is Sexist or Non-sexist.}
\newline

\noindent
In contrast to ISD, the LLMs in our experiments are biased towards generating ``Sexist'' and ``Non-sexist'' with the first letters capitalized.

\subsection{EDOS-B}
The task definition of EDOS-B is as follows:
\newline

\noindent
{\tt The goal of this task is to identify a category of a sexist text from Threat, Derogation, Animosity, or Prejudice.}
\newline

\noindent
The first letters are capitalized in the class labels with the same reason as in EDOS-A.
\citet{amb_icl} incorporated more detailed descriptions of the fine-grained class labels into their task definition, while we simply listed the class labels to make all the tasks be tested under the consistent setting.

\subsection{CLINC}
The task definition of CLINC is as follows:
\newline

\noindent
{\small {\tt The goal of this task is to identify a service domain and an intent given a user input. There are 10 domains: "auto\_and\_commute" "banking" "credit\_cards" "home" "kitchen\_and\_dining" "meta" "small\_talk" "travel" "utility" "work" . For each domain, there are 15 intents: "auto\_and\_commute"=[ "current\_location" "oil\_change\_when" "oil\_change\_how" "uber" "traffic" "tire\_pressure" "schedule\_maintenance" "gas" "mpg" "distance" "directions" "last\_maintenance" "gas\_type" "tire\_change" "jump\_start" ], "banking"=[ "freeze\_account" "routing" "pin\_change" "bill\_due" "pay\_bill" "account\_blocked" "interest\_rate" "min\_payment" "bill\_balance" "transfer" "order\_checks" "balance" "spending\_history" "transactions" "report\_fraud" ], "credit\_cards"=[ "replacement\_card\_duration" "expiration\_date" "damaged\_card" "improve\_credit\_score" "report\_lost\_card" "card\_declined" "credit\_limit\_change" "apr" "redeem\_rewards" "credit\_limit" "rewards\_balance" "application\_status" "credit\_score" "new\_card" "international\_fees" ], "home"=[ "what\_song" "play\_music" "todo\_list\_update" "reminder" "reminder\_update" "calendar\_update" "order\_status" "update\_playlist" "shopping\_list" "calendar" "next\_song" "order" "todo\_list" "shopping\_list\_update" "smart\_home" ], "kitchen\_and\_dining"=[ "food\_last" "confirm\_reservation" "how\_busy" "ingredients\_list" "calories" "nutrition\_info" "recipe" "restaurant\_reviews" "restaurant\_reservation" "meal\_suggestion" "restaurant\_suggestion" "cancel\_reservation" "ingredient\_substitution" "cook\_time" "accept\_reservations" ], "meta"=[ "change\_speed" "user\_name" "whisper\_mode" "yes" "change\_volume" "no" "change\_language" "repeat" "change\_accent" "cancel" "sync\_device" "change\_user\_name" "change\_ai\_name" "reset\_settings" "maybe" ], "small\_talk"=[ "who\_made\_you" "meaning\_of\_life" "who\_do\_you\_work\_for" "do\_you\_have\_pets" "what\_are\_your\_hobbies" "fun\_fact" "what\_is\_your\_name" "where\_are\_you\_from" "goodbye" "thank\_you" "greeting" "tell\_joke" "are\_you\_a\_bot" "how\_old\_are\_you" "what\_can\_i\_ask\_you" ], "travel"=[ "plug\_type" "travel\_notification" "translate" "flight\_status" "international\_visa" "timezone" "exchange\_rate" "travel\_suggestion" "travel\_alert" "vaccines" "lost\_luggage" "book\_flight" "book\_hotel" "carry\_on" "car\_rental" ], "utility"=[ "weather" "alarm" "date" "find\_phone" "share\_location" "timer" "make\_call" "calculator" "definition" "measurement\_conversion" "flip\_coin" "spelling" "time" "roll\_dice" "text" ], "work"=[ "pto\_request\_status" "next\_holiday" "insurance\_change" "insurance" "meeting\_schedule" "payday" "taxes" "income" "rollover\_401k" "pto\_balance" "pto\_request" "w2" "schedule\_meeting" "direct\_deposit" "pto\_used" ]. Then the output is like "domain: intent". If the input does not belong to any of the domains, the Answer is "oos".}}
\newline

\noindent
The class labels are grouped according to the domains.\footnote{\url{https://github.com/clinc/oos-eval/blob/master/data/domains.json}}
The output label is with the domain name, but in the evaluation, we only refer to the intent class.
The ``oos'' class is used to evaluate the detection of out-of-domain inputs~\citep{clinc}.

In our preliminary experiments, we have observed that simply using the LLMs' output strings leads very low recall of the ``oos'' class.
This is mainly because the LLMs are biased towards generating in-domain class labels affected by in-domain demonstrations, which is consistent with \citet{icl-bias-analysis-min} about LLMs' bias in ICL.
We then follow a threshold-based strategy in \citet{dnnc}; we replace an output of an in-domain class label with ``oos'' if its prediction probability is below a pre-defined threshold.
The threshold values are 0.6 for $k=1$, 0.7 for $k=2$, and 0.8 for $k=3,4,5$.

\subsection{SSENT}
The task definition of SSENT is as follows:
\newline

\noindent
{\tt The goal of this task is to copy the given hotel review by tagging sentiment-expressing phrases with the markup tags: <Positive></Positive> or <Negative></Negative>. Then the output is like "word1 <Positive>word2 word3</Positive> word4 <Negative>word5</Negative>". If there are not such tagged phrases, the Answer is Neutral.}
\newline

\noindent
For the output format, we followed \citet{karthik-formatting}.
The same task definition is shared in English and Spanish.

The structured sentiment analysis is a complex task in the original dataset~\citep{ssent}, where the task requires extracting tuples of (polar expression, subject, object) from an input text.
In our preliminary experiments, we observed that, without finetuning, it is not trivial how to handle this complicated task in few-shot ICL.

We then decided to only focus on the polar expression extraction task as a first step, and the subtask is considered to be a combination of sentiment classification and sequence labeling.
It is an interesting future direction to investigate end-to-end evaluation of such a complex task in ICL.

\subsection{XML-MT}
The task definition of XML-MT is as follows:
\newline

\noindent
{\tt The goal of this task is to translate an XML-tagged text from English to \underline{LANGUAGE} by preservin the XML structure. Both the input and output are like "word1 <tag-A>word2 word3</tag-A> word4 <tag-B>word5</tag-B>".}
\newline

\noindent
The placeholder {\tt LANGUAGE} is replaced with a language name ({\tt Finnish} or {\tt Japanese} in our experiments).

The test set has not been publicly released in the original dataset.\footnote{\url{https://github.com/salesforce/localization-xml-mt/tree/master/data}}
We sampled 500 examples from the original validation set for our validation set, and used the rest (1,500 examples) for our test set.

\section{Hallucination in LLMs' Prediction}
A concern about using the text generation models is hallucination~\citep{hal}.
In general, we cannot perfectly prevent the models from generating texts with hallucinations.
We do not discuss fact checking or semantics of the outputs, but instead solely focus on the output formats.

\subsection{ISD}
We did not observe any hallucinations on ISD in our experiments.
This is because we carefully selected the label strings as discussed in Appendix~\ref{app:data_isd}.

\subsection{EDOS-A}
We did not observe any hallucinations on EDOS-A in our experiments.
This is also because we carefully selected the label strings as discussed in Appendix~\ref{app:data_edos_a}.

\subsection{EDOS-B}
We observed a very small number of hallucinations on EDOS-B:
\begin{itemize}
    \item[1.] {\tt Sexual threat}
    \item[2.] {\tt Sexual Objectification}
    \item[3.] {\tt Objectification}
\end{itemize}
The case 1. is simply mapped to {\tt Threat}, and the cases 2. and 3. are mapped to {\tt Derogation} according to the description of the label in \citet{edos}.
However, the frequency is low; for example, the cases 2. and 3. are observed with 2 out of 486 examples on the validation set.

Note that we never observed such hallucinations when we used Flan-T5 XXL in Section~\ref{subsec:cross_lm}.
Therefore, this can be different across different LLMs.

\subsection{CLINC}
We observed only one hallucination case on CLINC, out of the 3,100 validation examples.
That case missed its domain label, only generating its intent label; in our evaluation pipeline, we automatically penalized such outputs that do not meet the defined output format.
However, as mentioned above, there was only one such case in our experiments.

\subsection{SSENT}
For SSENT, we have tried several different input-output formats, before we decided to use the format described in the task definition.
Specifically, the use of ``Neutral'' significantly reduced the chance to generate outputs with hallucinations.
We could not perfectly mitigate hallucinations on SSET, though.

However, we observed that most of the hallucinations can be trivially fixed with simple rules, by referring to the original inputs.
For example, the followings are typical cases:
\begin{itemize}
    \item original: don 't $\rightarrow$ model output: do n't
    \item original: weren 't $\rightarrow$ model output: were n't
    \item original: wasn 't $\rightarrow$ model output: was n't
\end{itemize}
We can see that the model outputs look more natural than the original tokens, where the LLM unexpectedly {\it fixed} the unnatural tokenization.
This is then more like a dataset issue.

Another type of hallucination was actually output formatting errors, such as missing a closing tag </Positive>.
For such hallucinations that are not trivially fixed, our evaluation pipeline automatically replaces the outputs with ``Neutral.''

\subsection{XML-MT}
For XML-MT, only the potential hallucination is XML tag formatting issue as in SSENT.
The formatting issue is rarely observed, but as in the original evaluation script,\footnote{\url{https://github.com/salesforce/localization-xml-mt/blob/master/scripts/evaluate.py}} we penalize such
predictions.

\section{Training of Reranking Model}

We use the T5X code base~\citep{t5x-paper}\footnote{\url{https://github.com/google-research/t5x}} for the reranking model.
Among the available checkpoints, we use mT5 XXL\footnote{\url{https://github.com/google-research/t5x/blob/main/docs/models.md\#mt5-checkpoints}} for all the experiments.
We finetune the original mT5 checkpoint to be used as a regression model~\citep{rank-t5}, once we create a set of training examples described in Section~\ref{subsec:training}.

We use the Adafactor optimizer~\citep{adafactor}, along with Z-loss regularization~\citep{z-loss}, a constant learning rate of 0.001, and a batch size of 256.
We evaluate checkpoints after every 100 updates of the model parameters; for the checkpoint selection, we use the reranking model to select demonstrations on the validation set of each dataset, and then select a checkpoint that leads to the best downstream metric with the LLMs' prediction.

To save the inference time of the LLMs on the validation sets, we pre-compute and cache all the possible $k$-shot predictions beforehand.
For each validation example, the number of the demonstration combinations is ${_n}{C}{_k}$, where demonstrations are ordered according to the original retrieval scores.
This caching strategy allows us to quickly check the LLMs' performance without repeating the inference multiple times.

\section{Effects of $\ell$ in Incremental Utility}
\label{app:inc_util}

\begin{table}[t]
\centering
\small{
\begin{tabular}{l|c|c|c|c}
\toprule
                   & $\ell=0.0$ & $\ell=0.5$ & $\ell=0.8$ & $\ell=1.0$  \\ \hline
$u_\mathrm{OP}^+$  & 46.91 & 46.80 & \textbf{48.80} & 46.98  \\
\bottomrule 
\end{tabular}
}
\caption{The effects of changing the value of $\ell$ in Equation~(\ref{eq:incremental_utility}), on the EDOS-B validation set.}
\label{tab:utility_tuning_1}
\end{table}

\begin{table}[t]
\centering
\small{
\begin{tabular}{l|c|c|c|c}
\toprule
                   & $\ell=0.0$ & $\ell=0.5$ & $\ell=0.8$ & $\ell=1.0$  \\ \hline
$u_\mathrm{OP}^+$  & 50.22 & 48.13 & \textbf{50.83} & 45.40  \\ \hdashline
$u_\mathrm{DM}^+$  & 49.11 & 49.38 & \textbf{49.82} & 45.18 \\
\bottomrule 
\end{tabular}
}
\caption{The effects of changing the value of $\ell$ in Equation~(\ref{eq:incremental_utility}), on the SSENT (es) validation set.}
\label{tab:utility_tuning_2}
\end{table}

In the beginning of Section~\ref{sec:results}, we have mentioned that $\ell=0.8$ in Equation~(\ref{eq:incremental_utility}) is recommended to implement the incremental utility function.
We report results of tuning the value of $\ell$ on a classification dataset, EDOS-B, and a non-classification dataset, SSENT (es).

Table~\ref{tab:utility_tuning_1} and Table~\ref{tab:utility_tuning_2} show the results on the validation sets.
Note that changing the values of $\ell$ does not affect the results with $u^{+}_\mathrm{DM}$ for the classification tasks, because the original reward values are binary (i.e., 0.0 or 1.0).
We can see that $\ell=0.8$ leads to the best scores, presumably because it takes a good balance between $\ell=0.0$ and $\ell=1.0$ as motivated in Section~\ref{subsec:inc_util}.

\section{Prompt Design for GPT-J}
\label{app:gptj}

GPT-J is a publicly available variant of the GPT models,\footnote{We used the transformers library for GPT-J: \url{https://huggingface.co/docs/transformers/model_doc/gptj}.} and it has been used in previous work~\citep{label-word-anchor}.
GPT-J is not an instruction-tuned model, and we design our prompt in a question-answering format:
\newline

\noindent
{\tt
\textbf{Text:} \underline{INPUT TEXT}\symbol{92}n \\
\textbf{Question:} which sexist category does the text express, Animosity, Derogation, Prejudice, or Threat?\symbol{92}n \\
\textbf{Answer:} the sexist category is
}
\newline

\noindent
The placeholder {\tt INPUT TEXT} is replaced with an actual input text, and this corresponds to the 0-shot inference.
Then, a prompt in 2-shot ICL is as follows:
\newline

\noindent
{\tt
\textbf{Text:} \underline{DEMONSTRATION INPUT 1}\symbol{92}n \\
\textbf{Question:} which sexist category does the text express, Animosity, Derogation, Prejudice, or Threat?\symbol{92}n \\
\textbf{Answer:} the sexist category is \underline{DEMONSTRATION OUTPUT 1}\symbol{92}n \\
\textbf{Text:} \underline{DEMONSTRATION INPUT 2}\symbol{92}n \\
\textbf{Question:} which sexist category does the text express, Animosity, Derogation, Prejudice, or Threat?\symbol{92}n \\
\textbf{Answer:} the sexist category is \underline{DEMONSTRATION OUTPUT 2}\symbol{92}n \\
\textbf{Text:} \underline{INPUT TEXT}\symbol{92}n \\
\textbf{Question:} which sexist category does the text express, Animosity, Derogation, Prejudice, or Threat?\symbol{92}n \\
\textbf{Answer:} the sexist category is
}
\newline

\noindent
Once the prompt is designed, we can seamlessly run the experiments with GPT-J.

\section{Compute Resources}

\paragraph{Flan-PaLM 2 (L)}
Readers may refer to \citet{palm2}.

\paragraph{Flan-T5 XXL}
The size of Flan-T5 XXL is 11B~\citep{flan-t5-paper}, and we used 64 v3 TPU chips to run inference with it.

\paragraph{GPT-J}
The size of GPT-J is 6B~\citep{gpt-j}, and we used one NVIDIA A100 GPU to run inference with it.

\paragraph{mT5 XXL for reranking}
The size of mT5 XXL is 13B~\citep{mt5paper}, and we used 64 v3 TPU chips to run both training and inference with it for reranking.

%% file: TAB-datasets_examples.tex
\begin{table*}[t]
\resizebox{\textwidth}{!}{
\centering
\begin{tabular}{l|l|l}
\toprule
Dataset & \multicolumn{1}{c|}{Input} & \multicolumn{1}{c}{Output} \\ \hline
CLINC & what expression would i use to say & travel: translate \\
      & i love you if i were an italian    & \\ \hline
SSENT & The food was very basic , but edible with & The food was <Negative>very basic</Negative> , but <Positive>edible \\
           & good bread , soup and desserts . & </Positive> with <Positive>good</Positive> bread , soup and desserts . \\ \hline
XML-MT & If you have permission to edit public templates, from  & Jos sinulla on oikeus muokata julkisia malleja, kirjoita Määritykset-valikon \\
       & Setup, enter <userinput>Email Templates</userinput>  & <parmname>Pikahaku</parmname>-kenttään <userinput>Sähköpostimallit \\
       & in the <parmname>Quick Find</parmname> box, then  & </userinput> ja valitse <uicontrol>Classic-sähköpostimallit</uicontrol>. \\
       & select <uicontrol>Classic Email Templates</uicontrol>. & \\
\bottomrule
\end{tabular}
}
\caption{Examples of input-output pairs in the training sets.}
\label{tab:task_examples}
\end{table*}